\documentclass[letterpaper, 10 pt, conference]{ieeeconf}  

\IEEEoverridecommandlockouts                              

\usepackage{graphics} 
\usepackage{epsfig} 
\usepackage{times} 
\usepackage{amsmath} 
\usepackage{amssymb}  
\let\labelindent\relax
\usepackage{enumitem}
\usepackage{hyperref}
\usepackage{xcolor}
\usepackage{color}
\usepackage{subfig}
\usepackage{multirow}
\usepackage{algorithm2e}
\usepackage{romannum}
\usepackage{float}
\usepackage{booktabs}
\usepackage[font=small,labelfont=bf]{caption}
\usepackage[normalem]{ulem}
\usepackage{url}
\usepackage{threeparttable}

\newcommand{\eg}{\emph{e.g.}}

\newcommand{\vect}[1]{\boldsymbol{#1}}

\title{\LARGE \bf
Transition Motion Planning for Multi-Limbed Vertical Climbing Robots Using Complementarity Constraints
}

\author{Jingwen Zhang$^{1}$, Xuan Lin$^{1}$, and Dennis W Hong$^{1}$
\thanks{$^{1}$Robotics and Mechanisms Laboratory (RoMeLa), Department of Mechanical and Aerospace Engineering, University of California Los Angeles, CA 90095.
{\tt\small zhjwzhang@g.ucla.edu, maynight@ucla.edu, dennishong@ucla.edu}}
}

\begin{document}

\maketitle
\thispagestyle{empty}
\pagestyle{empty}

\begin{abstract}

In order to achieve autonomous vertical wall climbing, the transition phase from the ground to the wall requires extra consideration inevitably. This paper focuses on the contact sequence planner to transition between flat terrain and vertical surfaces for multi-limbed climbing robots. To overcome the transition phase, it requires planning both multi-contact and contact wrenches simultaneously which makes it difficult. Instead of using a predetermined contact sequence, we consider various motions on different environment setups via modeling contact constraints and limb switchability as complementarity conditions. Two safety factors for toe sliding and motor over-torque are the main tuning parameters for different contact sequences. By solving as a nonlinear program (NLP), we can generate several feasible sequences of foot placements and contact forces to avoid failure cases. We verified feasibility with demonstrations on the hardware SiLVIA, a six-legged robot capable of vertically climbing between two walls by bracing itself in-between using only friction.

\end{abstract}

\section{INTRODUCTION}{\label{sec:intro}}

With many degrees of freedom, legged robots present an intriguing capacity for versatile motions in complex environments, such as disaster sites for search and rescue missions, and construction sites for maintenance. ZMP-based methods \cite{kajita2003biped} \cite{tedrake2015closed} provide a good starting point by assuming that the robot's contacts are co-planar. However, those scenes mentioned obviously involve uneven terrains and interactions with objects that require non-coplanar contacts. Many different locomotion strategies \cite{dai2016planning} \cite{winkler2017online} \cite{raibert1984hopping} \cite{nguyen2019optimized} are developed to handle uneven terrains and even obstacles with moderate height. However climbing is inevitable when faced tall obstacles, such as natural tunnels and man-made pipelines.

\begin{figure}[!ht]
    \centering
    \includegraphics[scale=0.45]{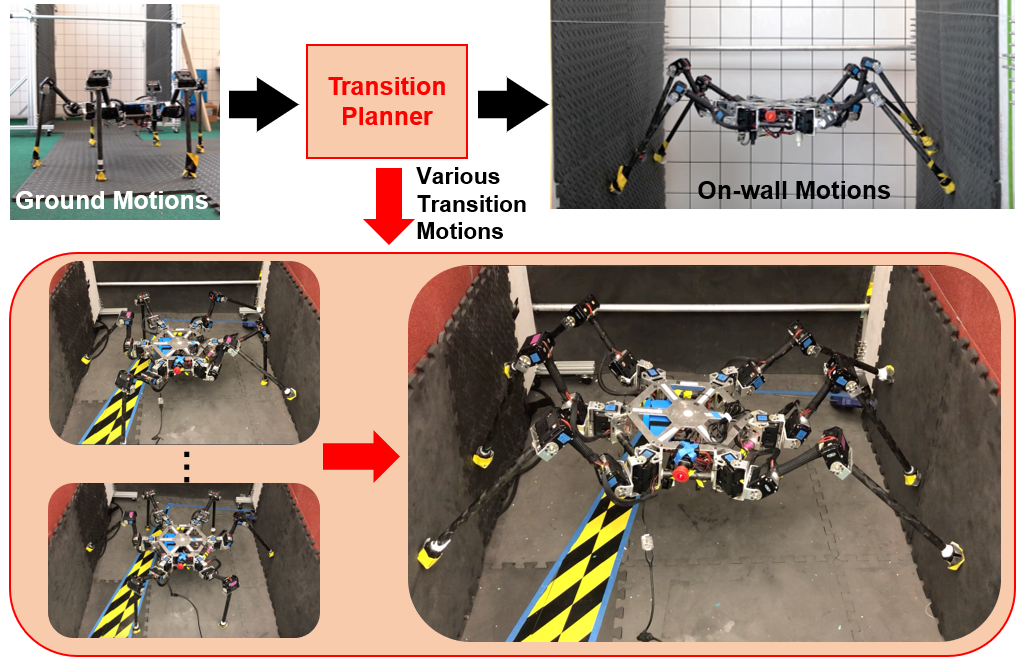}
    \caption{Proposed transition motion planner (red) is able to generate various motions in order to overcome the transition phase between grounds and walls.}
    \label{fig:concept}
\end{figure}

Many climbing robots \cite{kim2008smooth} \cite{spenko2008biologically} \cite{parness2017lemur} start by using gecko-type grippers or micro-spine grippers inspired by animals. Those designs make the system complex and task-specific. Planning climbing motions generally remains a challenging task for legged locomotion since it involves planning both multi-contact and contact wrenches simultaneously, especially when climbing more cluttered environments. The classical graph search method is used by \cite{bretl2006motion} in order to plan the motion on one single wall. Our previous work \cite{lin2019optimization} presents another approach which is based on optimization methods but the pre-determined contact sequence is required. Both methods assume that the initial condition of the robot on the wall is known and human operators are able to guarantee the initial setup at the beginning of the task. However, this assumption can be violated during an autonomous vertical wall climbing with minimal human intervention. Given the end posture of ground trajectory and initial posture of on-wall trajectory, the transition phase between ground and walls remains an open yet important question displayed in Fig. \ref{fig:concept}.

When planning climbing motions including the transition motion with contact, the intuitive approach is to pre-specify a contact sequence, and then optimize the trajectory for the fixed sequence. However, the number of possible sequences grows exponentially with the number of contact points which makes it difficult for operators to select an optimal one and challenges the discrete search methods in many cases \cite{tazaki2020survey}. Multi-contact planning by nonlinear programming (NLP) can avoid the combinatorial explosion of contact switching. \cite{winkler2018gait} presents a method to optimize over the discrete gait sequence using continuous variables through parameterization. By exploiting the complementarity condition between contact force and the distance to contact, \cite{posa2014direct} and \cite{dai2014whole} formulate it as a direct trajectory optimization problem for walking and jumping. In this paper we apply the same complementarity constraint to our motion planner and extend it to overcome the transition phase between the ground and walls.

This paper presents an NLP problem with complementarity constraints in order to explore different contact sequences during the transition phase given different setups, \eg, friction coefficients and motor torque limitations. Since the climbing motion for more complex, high degree-of-freedom robots is still quasi-static \cite{parness2017lemur}, the terms related to the derivative of momentum in the robot model are ignored. Besides the complementarity constraint between the contact force and the distance to contact, we also provide an optional complementarity constraint to capture the limb switchability, which describes whether the status (active/inactive) of one limb's contact should be switched, in order to mitigate the local minima issue in this problem. Safety factors for wall climbing are first proposed in our previous work \cite{lin2019optimization}. In this paper, we introduce the two safety factors, one for toe sliding and another for motor over-torque, into the algorithm and make them the main tuning parameters to reduce the gap between our planner and the hardware experiments, \eg, inaccurate friction coefficients and motor heating problems. Although with strong nonlinearity from complementarity constraints, the sparsity of the resulting problem enables us to get efficient (locally optimal) solutions with solvers such as SNOPT \cite{gill2005snopt}. We implement the resulting solutions directly on our hardware SiLVIA and verify the feasibility with demonstrations on several different experiments. To the best of our knowledge, this is the first hexapod robot who can overcome the transition phase between the ground and walls in the real world with regular end effectors that only use frictional forces.

This paper is organized as follows. In Section \ref{sec:approach} we describe the simple model and introduce variations of our constraints in the formulation. In Section \ref{sec:results}, we show our results on SiLVIA performing a variety of transition motions between the wall and the ground. We conclude the discussion in Section \ref{sec:conclusions}.

\section{APPROACH}{\label{sec:approach}}
When a legged robot is moving from supporting by the ground reaction force to bracing itself between walls with only frictional forces, static equilibrium constraints are typically used with the assumption of quasi-static movements in order to ensure the stability. Kinematic reachability and dynamic constraints like forcing contact forces inside of a friction cone are important as well for feasible planning.

\subsection{Static Equilibrium Constraint}
Generally, legged robots with $n$ joints will be treated as an underactuated system with a total of $n+6$ degrees of freedom (DOF) since the floating base, which is often the body of a robot, cannot be controlled directly. The additional DOF will be determined by wrenches (force/torque) applied on the robot including contact wrenches and gravity. According to the centroidal dynamics \cite{orin2013centroidal}, we can represent them as linear and angular momentum at the center of mass (COM). Climbing on the wall only with the friction generated by regular end-effectors, like the point-contact toe, requires large motor torques which is prone to causing over-torque. The additional acceleration (linear/angular) of the body will burden motors significantly. For safety consideration, quasi-static movements where the derivative of linear and angular momentum can be negligible are presumed for our applications. With this assumption, the robot is subject to the static equilibrium constraint for every round $j$ where $j = 1, \ldots, M$ and $M$ is the total rounds to be planned.
\begin{subequations}\label{const:static}
\begin{align}
    \sum^{N}_{i=1} \vect{f}_{i,j} + m\vect{g} = \vect{0} & \\
    \sum^{N}_{i=1} (\vect{p}_{i,j} - \vect{c}_{j}) \times \vect{f}_{i,j}  + \vect{\lambda}_{i,j} = \vect{0}
\end{align}
\end{subequations}
where $m$ is the total mass of the robot, $\vect{c}_{j} \in \mathbb{R}^3$ is the COM position, $\vect{p}_{i,j} \in \mathbb{R}^3$ and $(\vect{f}_{i,j}, \vect{\lambda}_{i,j}) \in \mathbb{R}^6$ denote the position and contact wrench for the $i^{th}$ leg ($N$ is the total number of limbs), with respect to the world frame. The contact wrench will be kept at zero if the contact is inactive.

\subsection{Reachability}\label{subsec:Reachability}
Given a legged robot, the design will put a variety of constraints on the motion of the robot geometrically. For every round $j$, given the COM position $\vect{c}_{j} \in \mathbb{R}^3$ and the body orientation $\vect{\Theta}_{j} \in \mathbb{R}^3$, we can specify the limb reachability as follows:
\begin{equation}\label{const:reachability}
    \vect{p}_{i,j} \in \mathcal{R}(\vect{c}_{j}, \vect{\Theta}_{j})
\end{equation}
where the set $\mathcal{R}(\vect{c}_{j}, \vect{\Theta}_{j})$ can be expressed explicitly with either analytical expressions by introducing additional decision variables joint angles $\vect{\theta}_{i, j}$ for each leg $i$, or approximated ranges. With full kinematics model, we can express the constraint \eqref{const:reachability} exactly as follows:
\begin{subequations}\label{eq:reach}
\begin{align}
    \vect{p}_{i,j} = \phi_{FK} (\vect{\theta}_{i, j}) \\
    \vect{\theta}_{min} \leq \vect{\theta}_{i, j} \leq \vect{\theta}_{max}
\end{align}
\end{subequations}
where $\phi_{FK}(\cdot)$ and $\vect{J}(\vect{\theta}_{i, j})$ denote the forward kinematics and Jacobian matrix for each leg, $\vect{\theta}_{min}$ and $\vect{\theta}_{max}$ describe the physical joint limitations. Although full kinematics model enables rich constraints including the position/orientation of every end-effector and explicit joint limitation, it will increase the computational expense with additional decision variables and high nonlinearility introduced by forward kinematics and Jacobian matrix. With proper approximated ranges, we can still capture the physical meanings behind those constraints although it will be more conservative. For the reachability \eqref{const:reachability}, we can approximate the leg reachable workspace as a sphere around the nominal posture which can be expressed as:
\begin{equation}\label{const:estimated reach}
    \left\lVert \vect{p}_{i,j} - \vect{c}_{j} - \vect{R}(\vect{\Theta}_{ j})\vect{v} \right\rVert_2 \leq \Delta_{FK}
\end{equation} 
where $\vect{v}$ denotes the offset from COM to the housing location of legs on the body, $\vect{R}(\vect{\Theta}_{ j})$ is the body rotation matrix represented by Euler angles, and $\Delta_{FK}$ is the radius of the approximated sephere. Furthermore, the step size of both the body and limbs is considered to ensure the reachability between rounds as follows:
\begin{subequations}\label{const:stepsiz}
\begin{align}
|\Delta \vect{c}| \leq \Delta \vect{c}_{max}  \\
|\Delta \vect{\Theta}| \leq \Delta \vect{\Theta}_{max} \\
|\Delta \vect{p}_{i}| \leq \Delta \vect{p}_{max} 
\end{align}
\end{subequations}
where $\Delta \vect{c}_{max}$, $\Delta \vect{\Theta}_{max}$ and $\Delta \vect{p}_{max}$ denote maximum stepsizes for the body's linear translation, rotation and maximum limb stride length respectively.

\subsection{Contact Wrench}
\begin{figure}[!ht]
    \centering
    \includegraphics[scale=0.5]{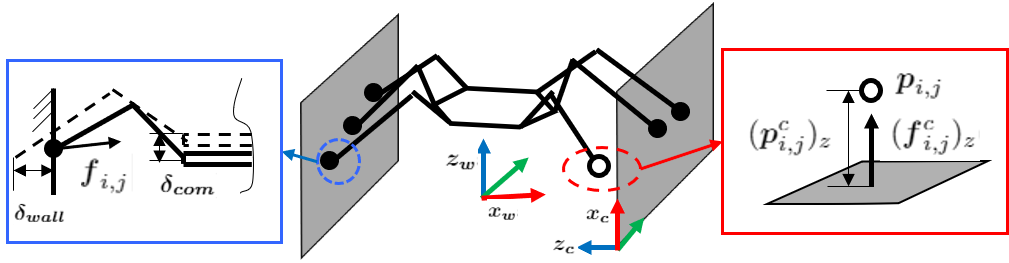}
    \caption{Illustration of the world frame $\{w\}$ and the local contact frame $\{c\}$. Solid dots indicate toes on the wall while the hollow one indicates the swing leg. Inside the red box is the complementarity condition between contact distance $(\vect{p}^{c}_{i, j})_z$ and the normal contact force $(\vect{f}^{c}_{i, j})_z$. Inside the blue box is the limb compliance for indirect force control of SiLVIA.}
    \label{fig:illustration}
\end{figure}

Since wall climbing is a high-risk task, being conservative on the choice of contact wrenches is necessary for safety of both the robot and the operator. Two safety factors are defined as follows in order to deal with the friction coefficient $\mu$ which is hard to be measured precisely and the motor torque limit $\tau_{max}$ which can degenerate by heating. More details can be seen in our previous work \cite{lin2019optimization}.
\begin{align}
S_{\mu}&=\mu/\mu_{c} \\
S_{\tau}&=\tau_{max}/\tau_{c}
\label{Eqn:friction_required}
\end{align}
where $\mu_{c}$ and $\tau_{c}$ denote the critical values for safety. For our algorithm, critical values are not determined experimentally. Instead, we accept $S_{\mu}$ and $S_{\tau}$ as two main tuning parameters and being conservative can be achieved via keeping $S_{\mu} \geq 1$ and $S_{\tau} \geq 1$. Similarly, the constraint on contact wrench can include two parts, one from the motor torque limits and another from the friction cone.
\begin{equation}\label{const:wrench}
    \vect{f}_{i, j} \in \mathcal{F}_{\tau}(\tau_{max}, \vect{J}(\vect{\theta}_{i, j}), S_{\tau}) \cap \mathcal{F}_{\mu}(\mu, S_{\mu})
\end{equation}
where the contact torque is not included for consideration since most legged robots are assumed point contact generally (multiple point contacts at corners of the sole are used for humanoids). If the torque can be generated by the end-effector and is controllable, it is easy to modify the contraint \eqref{const:wrench}. Actually, the contact torque is very important for stability of a few contact sequences which we will see in Section \ref{sec:results}. The friction cone can be expressed as follows:
\begin{equation}
    \mathcal{F}_{\mu}(\mu, S_{\mu}) = \left\{\vect{f}_{i, j} \mid \left\lVert (\vect{f}^{c}_{i, j})_{x,y} \right\lVert^2_2 \leq \mu(\vect{f}^{c}_{i, j})_z / S_{\mu} \right\}
\end{equation}
where $\vect{f}^{c}_{i, j}$ denotes the contact force for the $i^{th}$ leg in the $j^{th}$ round with respect to the contact frame as shown in Fig. \ref{fig:illustration}. $(\cdot)_{x,y}$ extracts the $x$ and $y$ components while $(\cdot)_{z}$ is the $z$ component of the vector. In terms of the set $\mathcal{F}_{\tau}(\tau_{max}, \vect{J}(\vect{\theta}_{i, j}), S_{\tau})$, two explicit options are provided similar to Section \ref{subsec:Reachability}. Based on full kinematics model, we can express it as:
\begin{subequations}\label{eq:force limit}
\begin{align}
    \vect{\tau}_{i,j} = \vect{J}(\vect{\theta}_{i, j})^\top \vect{f}_{i,j} \label{eq:jac}\\
    \vect{\tau}_{i,j} \leq \vect{\tau}_{max} / S_{\tau}
\end{align}
\end{subequations}
where $\vect{\tau}_{i,j}$ denotes required torques of motors on the $i^{th}$ leg. We can also assume all actuators have the same torque limit $\tau_{max}$ which is a scalar value so that we require the maximum of joint torques not to exceed the relaxed limit during operation,
\begin{equation}\label{eq:torque limit}
    \left\lVert \vect{\tau}_{i,j} \right\rVert_{\infty} \leq \tau_{max} / S_{\tau}
\end{equation}

With the inequality $\left\lVert \vect{\tau}_{i,j} \right\rVert_{\infty} \leq \left\lVert \vect{J}(\vect{\theta}_{i, j})^\top \right\rVert_{\infty} \left\lVert \vect{f}_{i,j} \right\rVert_{\infty} $ from Eq.~\eqref{eq:jac}, we can set the right-hand side of Eq.~\eqref{eq:torque limit} as a conservative upper bound for $\vect{J}(\vect{\theta}_{i, j})^\top_{\infty} \left\lVert \vect{f}_{i,j} \right\rVert_{\infty}$, and the following relation can be obtained:
\begin{equation}\label{const:estimated force}
    \left\lVert \vect{f}_{i,j} \right\rVert_{\infty} \leq \frac{\tau_{max}}{S_{\tau} \underset{\vect{\theta}_{i, j}}{\textit{max}} \left\lVert \vect{J}(\vect{\theta}_{i, j})^\top \right\rVert_{\infty}} 
\end{equation}
and for a robot, we can treat the value of $\underset{}{\textit{max}} \left\lVert \vect{J}(\vect{\theta}_{i, j})^\top \right\rVert_{\infty} $ as a constant by searching all possible $\vect{\theta}_{i, j}$ offline.

\subsection{Unscheduled Contact Sequence and Limb Switchability}
As discussed previously in Section \ref{sec:intro}, we apply the same idea of \cite{posa2014direct} exploiting the complementarity condition between the contact distance and the contact force to our motion planning algorithm. As shown in Fig. \ref{fig:illustration}, the contact force can be generated only when the contact distance is zero. Therefore, either the contact distance $(\vect{p}^{c}_{i, j})_z$ or the normal contact force $(\vect{f}^{c}_{i, j})_z$ will be zero which can be express as:
\begin{subequations}\label{const:comple}
\begin{align}
   (\vect{f}^{c}_{i, j})_z \ge 0, \quad (\vect{p}^{c}_{i, j})_z \ge 0 \\ (\vect{f}^{c}_{i, j})_z (\vect{p}^{c}_{i, j})_z = 0 
\end{align}
\end{subequations}

It is noted that the complementarity condition between the contact distance and the contact torque which can be expressed as $\lVert {\lambda}^{c}_{i, j} \rVert^2 (\vect{p}^{c}_{i, j})_z = 0 $ is ignored here for the point contact assumption. In this way, we can use only continuous variables in the optimization problem and search over all possible contact sequences at once.

Furthermore, we also provide an optional complentarity constraint for limb switchability. If we use the same leg for supporting the weight during the transition phase for too many rounds which means a long period, it is likely for motors of that leg to over torque due to heating. In order to reduce this risk, we can force the motion planner to switch supporting legs between rounds, expressed as follows:
\begin{equation}\label{const:switchability}
   \text{for $j = 2, \dots, M$}, \quad (\vect{f}^{c}_{i, j-1})_z(\vect{f}^{c}_{i, j})_z = 0 
\end{equation}
If one leg is used to generate nonzero contact force in the prior round, the complentraity condition will force the normal contact force of the same leg for the current round to be zero which means this leg will not be one of supporting legs. The observation here is that it will also sacrifice some richness in the possible transition motions. Without the constraint \eqref{const:switchability}, we can generate more feasible contact sequences which can be seen from our results in Section \ref{sec:results}.

\subsection{Complete NLP formulation}
With all constraints described previously, we can choose the decision variables  for the $j^{th}$ round as:
\begin{equation}
\mathit{\Gamma} = \left\{\vect{p}_{i, j}, \vect{f}_{i, j}, \vect{c}_{j}, \vect{\Theta}_{j}\right\}
\end{equation}

If the explicit form of constraints on reachability and the motor torque limits is chosen as Eq.~\eqref{eq:reach} and Eq.~\eqref{eq:force limit}, additional decision variable $\vect{\theta}_{i, j}$ will be required. Similarly, $\vect{\lambda}_{i,j}$ can be introduced if consider contact torques explicitly.

In terms of the cost function, a terminal cost and intermediate costs are included. To prepare the robot ready for wall-climbing during transition motion planning, we command the robot to reach the initial posture of the following climbing trajectory which is our terminal condition. Intermediate costs are composed of stepsize penalty and the $L_1$ norm of contact force. Unlike the $L_2$ norm of contact force which would encourage evenly distributed forces, the $L_1$ norm would favor sparse solutions which is what we expect from complementarity constraints. The full NLP formulation can be expressed as:

$\begin{aligned}
& \begin{aligned}
    \underset{\Gamma}{\textbf{minimize}} \quad & \sum^{N}_{i=1} |\vect{p}_{i, M} - \vect{p}_{i, d}|^2_{Q_p} + \sum^{M}_{j=2} (|\Delta \vect{c}_j|^2_{Q_c}  \\
    & + |\Delta \vect{\Theta}_j|^2_{Q_{\Theta}} + \sum^{N}_{i=1} |\Delta \vect{p}_{i,j}|^2_{Q_{\Delta}} + |\vect{f}_{i,j}|^2_{L_1} ) 
\end{aligned} \\
& \text{subject to \eqref{const:static}, \eqref{const:reachability}, \eqref{const:stepsiz}, \eqref{const:wrench}, \eqref{const:comple}, \eqref{const:switchability} (optional)}\\
\\
\end{aligned}$

where $|\cdot|^2_Q$ is the abbreviation for the quadratic cost with the weight matrix as $Q \geq 0$, $\vect{p}_{i, M}$ and $\vect{p}_{i, d}$ denote the position at the last round $M$ and desired position for the $i^{th}$ leg respectively. Note that we could also add the contact position assignment as a constraint represented by $\vect{A}\vect{p}_{i, j} \leq \vect{b}$ where $\vect{A}$ and $\vect{b}$ describe the convex hull of feasible contact regions.

\subsection{Indirect Force Control for SiLVIA}
After solving our transition motion planner with SNOPT, the robot is able to accomplish the transtion task from the ground to the wall round by round by commanding the optimized leg positions and contact forces. If the robot has force feedback on the end-effector of each leg, direct force control would work easily. Unfortunately, our robot SiLVIA is position controlled. For multi-limbed robots like SiLVIA, contact forces are statically indeterminate if more than 3 contacts are active. In \cite{lin2018multi}, the virtual penetration into the wall $\vect{\delta_{wall}}$ and the body deformation $\vect{\delta_{com}}$ as shown in Fig. \ref{fig:illustration} are considered to determine contact forces indirectly. The optimized contact force $\vect{f}_{i,j}$ from our transition motion planner can be treated as the spring force using the Virtual Joint Method (VJM) \cite{pashkevich2011enhanced}. In order to achieve the contact force at the $i^{th}$ leg for the $j^{th}$ round, we need to solve a feasibility problem as follows:
\begin{subequations}\label{eq:delta_wall}
\begin{align}
    \textbf{find}  \quad & \vect{\delta_{wall}}, \vect{\delta_{com}}
    \tag{\ref{eq:delta_wall}}  \\
& \vect{f}_{i,j} = \vect{K}_{i,j} (\vect{\delta_{wall}} - \vect{\delta_{com}}) & \\
& \vect{K}_{i,j} = (\vect{J}(\vect{\theta}_{i, j}) \vect{k}^{-1} \vect{J}(\vect{\theta}_{i, j})^\top)^{-1}
\end{align}
\end{subequations}
where $\vect{K}_{i,j}$ is the stiffness matrix of the $n$ DOF leg, $\vect{k}$ is the diagonal matrix with the virtual spring coefficients $[k_1, k_2, \ldots, k_n]$ of the position controller motors. Since $\vect{\theta}_{i,j}$ can be find after transition motion planner, only $\vect{\delta_{wall}}$ and $\vect{\delta_{com}}$ are unknown. Problem \eqref{eq:delta_wall} can be solved efficiently. Therefore indirect force control can be achieved by adding $\vect{\delta_{wall}}$ to optimized leg positions $\vect{p}_{i,j}$ which is what we have done for the following experiments.
\section{RESULTS}{\label{sec:results}}
In this section, we presents several different scenarios about transition between the wall and the ground, including the parallel wall, the parallel wall with steps or inclines, and the circular wall. Without losing generality, we also try to explore walking gaits on the ground. With the proposed transition motion planner, we choose Eq. \eqref{const:estimated reach} and Eq. \eqref{const:estimated force} to represent the constraint \eqref{const:reachability} and \eqref{const:wrench} respectively in order to reduce the computational cost. The constant $\underset{}{\textit{max}} \left\lVert \vect{J}(\vect{\theta}_{i, j})^\top \right\rVert_{\infty} $ is determined offline as 963.5 mm. Results are implemented on the six-legged robot SiLVIA weighted around 10 kg, each leg of which has 3 DOF. A pair of Dynamixel MX-106 motors is used for each joint which can provide a maximum torque at 27 Nm. For the resulting contact sequence, B-spline is used for interpolation and we can generate the trajectory for the independent joint PID controller. All hardware demonstrations can be viewed in the accompanied video.

\subsection{The Parallel Wall}\label{subsec:parallel wall}
\begin{figure}[!ht]
    \centering
    \includegraphics[scale=0.5]{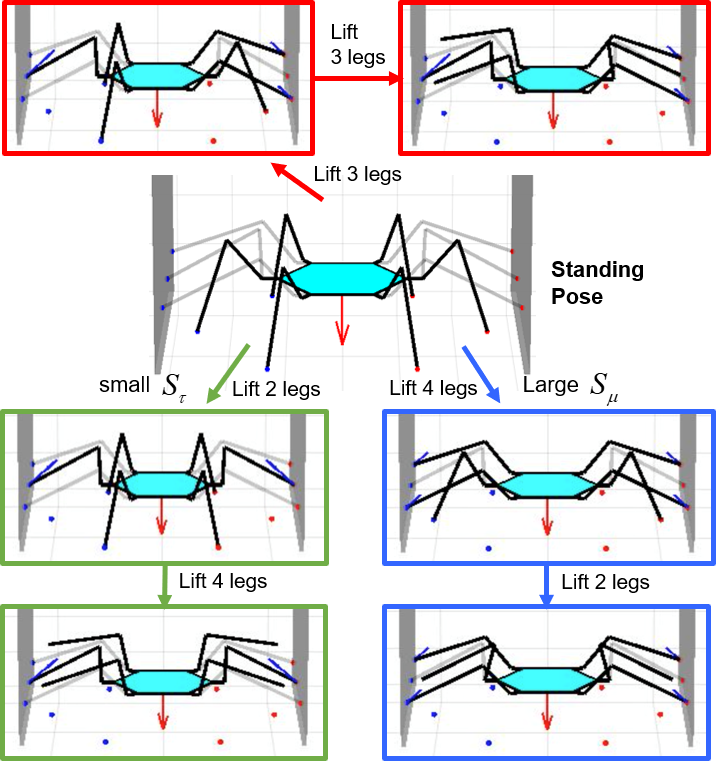}
    \caption{Simulation visualization of 3 different resulting contact sequences. The red arrow inside each graph denotes the gravity while blue arrows indicate contact forces. The translucent model is the desired configuration for transition. Red path represents 3-3 contact sequence. Green path represents 2-4 contact sequence with small $S_{\tau}$. Blue path represents 4-2 contact sequence with large $S_{\mu}$.}
    \label{fig:parallel wall}
\end{figure}

In this scenario, we let the robot stand between two parallel walls at a distance of 1230 mm. Initially, the robot stands on the ground with the body height as 210 mm. The desired configuration on the wall is the starting point of the climbing trajectory from our previous work \cite{lin2019optimization} indicated in Fig. \ref{fig:parallel wall} as the translucent configuration.

 In the hardware experiment, the walls are covered by rubber pads and the robot toes are covered by anti-slip tapes, which gives a frictional coefficient $\mu$ around 1. With the nominal values for safety factors ($S_{\tau} = 1.8$, $S_{\mu} = 1.1$), the tripod gait (3-3 contact sequence) which lifts the front and rear legs on one side, the middle leg on the other side firstly, and then lifts the remaining 3 legs, is generated and validated in the robot as seen in Fig. \ref{fig:concept}. In the following, the two safety factors are tuned to simulate different experimental setups although the real environment for hardware demonstrations is kept the same. For the planner, we make the wall surface more slippery by increasing the value of $S_{\mu}$. When $S_{\mu} \approx 1.46$, the resulting solution shows the robot would lift two front and two rear legs for the first round, and then lift two remaining middle legs to the wall (4-2 contact sequence in Fig. \ref{fig:parallel wall}). We also increase the motor torque limit by decreasing $S_{\tau}$ to pretend that the robot has stronger motors. 2-4 contact sequence is generated when $S_{\tau}$ becomes around 0.85. Actually, this is an aggressive setting for our robot since the torque density needs to be $1/0.85$ times larger in order to execute the resulting solution successfully in the hardware. It is noted that with smaller $S_{\tau}$, both 3-3 contact sequence and 4-2 contact sequence are also feasible local minimas of the NLP planner. With a different initial guess, the planner would converge to a different local minima.
 
\begin{figure}[!ht]
    \centering
    \includegraphics[scale=0.5]{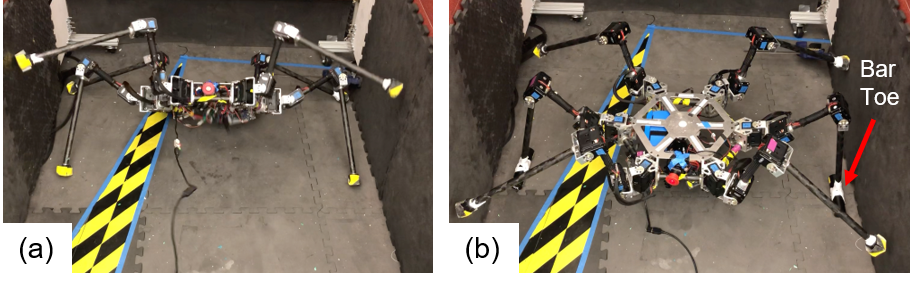}
    \caption{(a) Failure with point-contact toes for the two middle legs. (b) Success with bar toes for the two middle legs.}
    \label{fig:bar toe}
\end{figure}

 The interesting observation in the hardware experiment is that the robot tips over for those two contact sequences (2-4 and 4-2). In terms of the 4-2 contact sequence, falling down happens when lifting those 4 legs as indicated in Fig. \ref{fig:bar toe}. The reason is that when the robot only has 2 legs generating contact forces, they need to be perfectly aligned in order to avoid any torques applied on the body. The better strategy is to include the contact torque in the planner. However, SiLVIA only has 3 DOF each leg which means the contact torque on the end-effector cannot be controlled directly. We simply replace the point-contact toe with a bar to be robust passively and it performs the 4-2 contact sequence successfully. For the 2-4 contact sequence, the robot suffers from not only the same issue as noted above but also the over-torque problems for the two legs on the wall.
 
\begin{figure}[!ht]
    \centering
    \includegraphics[scale=0.09]{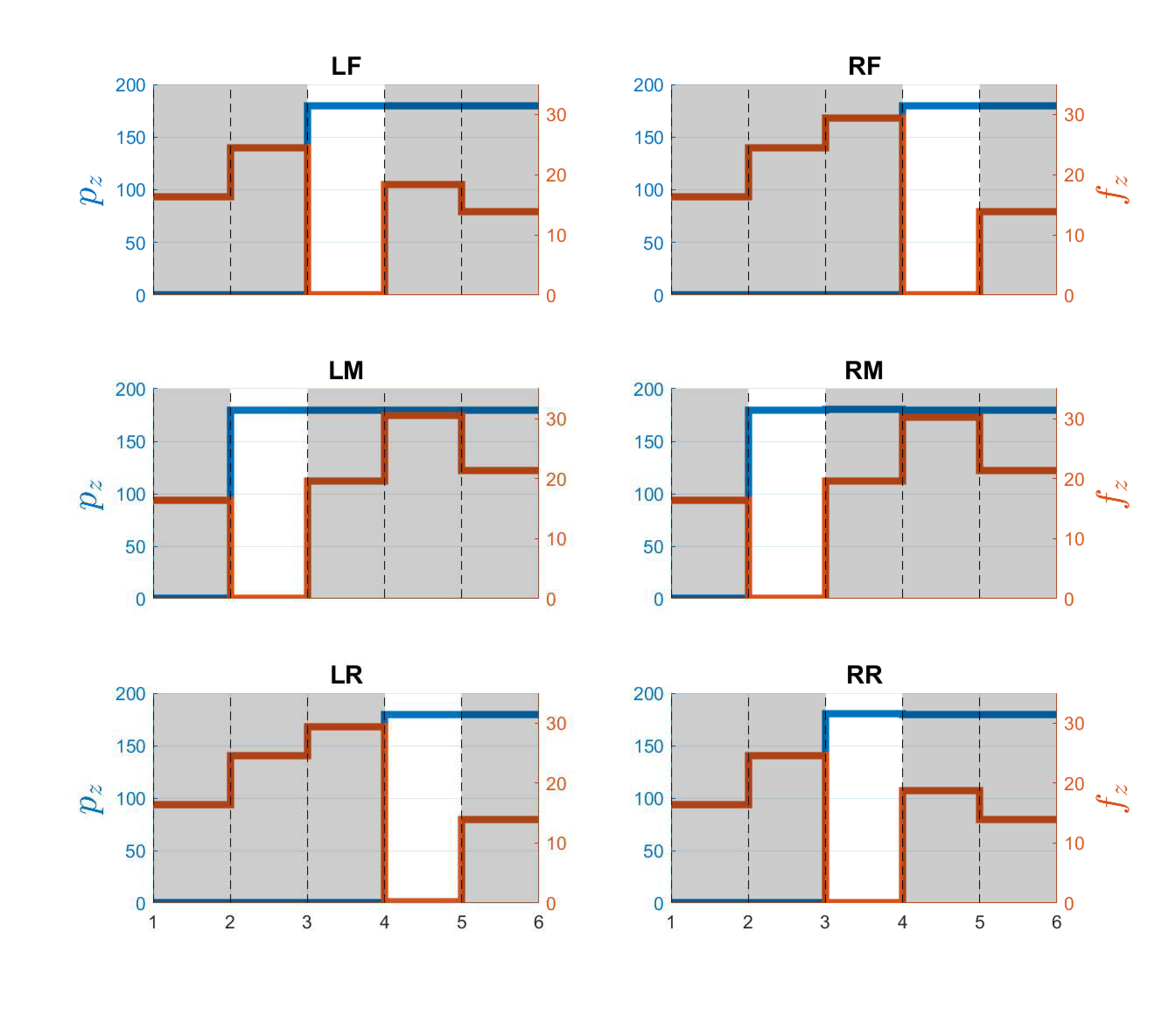}
    \caption{The resulting z-component of the contact forces and toe positions with respect to the world frame without limb switchability constraint. Shaded areas indicate the contact is active for the leg.}
    \label{fig:2-2-2 contact sequence}
\end{figure}

Furthermore, we ignore the limb switchability constraint \eqref{const:switchability} in order to explore more possible contact sequences. A new contact sequence (2-2-2 contact sequence) is generated by our planner. The robot would lift the left middle leg (LM) and the right middle leg (RM) initially, then lift the left front leg (LF) and the right rear leg (RR) with 4 legs for supporting (2 on the ground and 2 on the wall), and finally lift the remaining two legs to the wall. From Fig. \ref{fig:2-2-2 contact sequence}, the complementarity constraint \eqref{const:comple} is always satisfied. For example, the contact distance of the leg RM at the round 1 is 0 while the contact force is 0 for the round 2. Note that the leg RM is already on the wall after the round 2 (the desired height of the leg on the wall is 180 mm). For rounds 3, 4 and 5, the contact distance of the leg RM is 0 with respect to the contact frame which still keeps the complementarity condition. The hardware demonstration shows that the 2-2-2 contact sequence provides stronger stability of the transition motion since the robot always keeps at least 4 contact points. 

\subsection{The Parallel Wall with Steps}
\begin{figure}[!ht]
    \centering
    \includegraphics[scale=0.4]{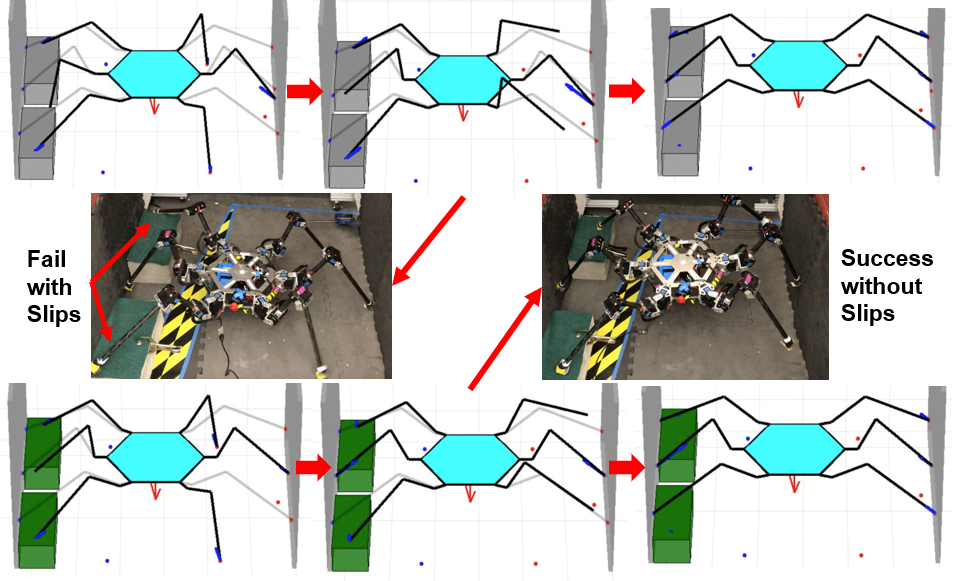}
    \caption{The upper row is the 3-3-3 contact sequence for the parallel wall with steps while the bottom row is the 3-1-2-2 contact sequence. The middle row is the screenshots of hardware experiments when implementing these two contact sequences. (Green represents a surface with smaller friction coefficient)}
    \label{fig:with steps}
\end{figure}

This scenario focuses on investigating whether our planner can generate the contact sequence taking advantage of objects presented in order to overcome the transition phase. We put two bricks next to the left wall on the ground and intuitively the robot would be able to step on the brick for the intermediate round. To take advantage of the brick, we decrease the value of $\Delta_{FK}$ in the constraint \eqref{const:estimated reach} for the legs on the left side until the leg is unable to reach the desired positions on the wall in one step. The 3-3-3 contact sequence is generated which puts LF and LR on the brick and RM on the right wall, then moves RF, RR and LM on the wall, finally moves LF, LR and RM again to the desired positions on the wall, as shown in Fig. \ref{fig:with steps}. However, in the hardware experiment, the robot tips over when it is trying to lift 3 legs based on the supporting from 2 legs on the brick and 1 leg on the wall. The two legs on the brick (LF and LR) slip causing the failure. The reason is that the friction coefficient between the brick and the toe is much smaller than the one between the toe and the wall. By realizing this, we correct the setting but the problem becomes infeasible. Then we add one more round to enrich the possible contact sequence that the planner can search over. From the bottom row of Fig. \ref{fig:with steps}. We can see that for the second round, the robot would not lift 3 legs together as what the robot would do in the 3-3-3 contact sequence. Instead, it is divided into two rounds. The robot would lift the leg LM to create 4 contact points and then lift the right two legs (RF and RR) to avoid too large horizontal forces required for the two legs on the brick. The robot successfully overcomes the transition phase with steps by performing the 3-1-2-2 contact sequence.

\subsection{Others}
\begin{figure}[!ht]
    \centering
    \includegraphics[scale=0.4]{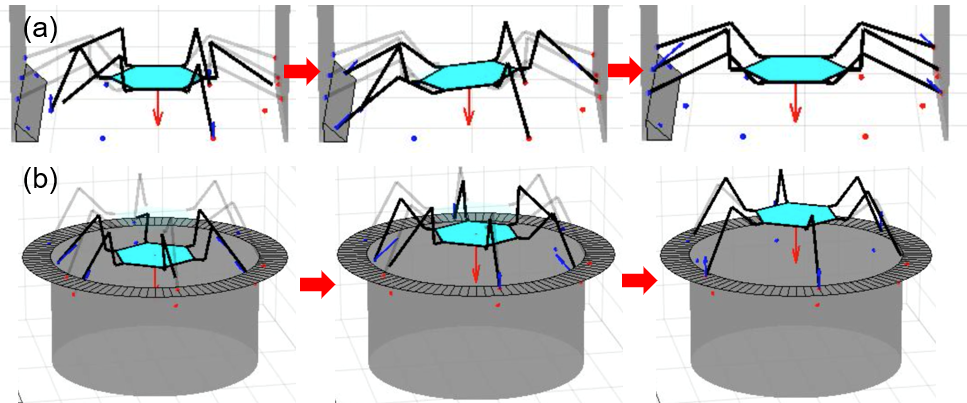}
    \caption{(a) The transition motion with the parallel wall with an incline. (b) The transition motion from a tube to the flat platform outside the tube.}
    \label{fig:others}
\end{figure}

Besides the scenarios described previously, we also simulate two more cases. The first one is that we replace the two bricks with an incline to investigate the capability of changing the body orientation. We use $L_2$ norm of contact force in the cost function to encourage forces to be distributed more evenly which is also one possible incentive to change the body orientation. We also reduce the weight $Q_{\Theta}$ of the body orientation stepsize. The second one is to investigate how to transit from the wall to the ground. Obviously, reversing the order of our previous contact sequence would work since the robot is under the static equilibrium for every round. However, when you want to climb out of a circular wall or a tube, the situation is totally different since the flat surface the robot can place the leg is around the tube externally. Both simulation results are visualized in Fig. \ref{fig:others}.

\begin{figure}[!ht]
    \centering
    \includegraphics[scale=0.4]{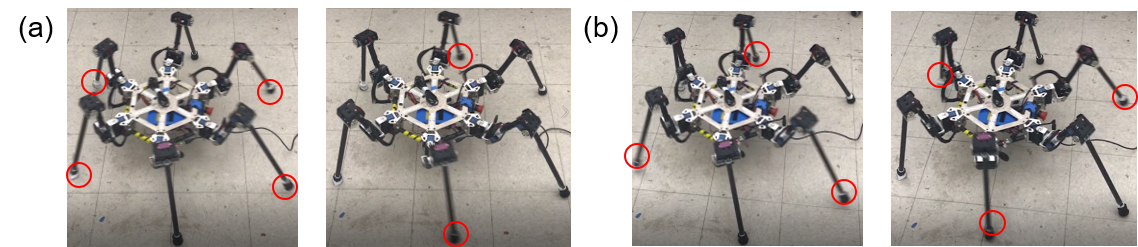}
    \caption{Red circles highlight the moving legs. (a) Screenshots for the repeated 2-4 gait (b) Screenshots for the repeated 3-3 gait.}
    \label{fig:walking}
\end{figure}

Without losing generality, we also apply our transition motion planner for ground walking. With different initial guesses, our planner converges to different local minimas and generates the standard tripod gait (repeated 3-3 contact sequence actually) and repeated 2-4 gait. The hardware demonstration is shown in Fig. \ref{fig:walking}.

\subsection{Running Time}
Generally, a big concern about NLP, especially with complementarity constraints, is the running time. Since only a small number of rounds $M$ (generally $\leq 5$) required for the transition motion, our MATLAB code using SNOPT as the solver usually generates motions described previously within 10 secs. Also, NLP is very sensitive to the initial guess. A warm start that ignores highly nonlinear complementarity constraints can always help find a better initial guess leading to less computational time. Currently, for the parallel wall task, the average running time varies from 1.054s to 3.083s. These numbers are reported by running on an Intel Core i7-6500U CPU. We believe the implementation with the C++ code can accelerate the planner and make it run online which is also one of our future works.
\section{DISCUSSION AND CONCLUSION}{\label{sec:conclusions}}
Although we propose that the contact wrench of the end-effector can be optimized through our motion planner, multi-limbed robots, unlike humanoid robots, generally are not designed with enough DOF to fully control the contact wrench (in most cases, the contact torque cannot be controlled directly). However, as we have seen from Section \ref{subsec:parallel wall}, the contact torque can play a big role in improving the robustness of the resulting transition motions. Introduction of potential coupling relation between controllable forces and uncontrollable torques remains an open question. For our proposed method, the planning round $M$ needs to be determined based on human experience. Investigation on planning $M$ simultaneously would be one of the future works. Quantitative analysis based on appropriate performance metrics, \eg, success rate, running speed and the resulting joint torque can be used to determine the optimal settings. Our proposed method assumes prior knowledge of estimating the friction coefficient. Fortunately, recent developments on image processing and learning technique make this prior condition possible with on-board vision systems.


In this paper we present a transition motion planner for multi-limbed vertical climbing robots to overcome the transition phase between the ground and walls. Complementarity constraints are used to generate unscheduled contact sequences and describe how to switch supporting limbs. The optional complementary constraint on limb switchability helps mitigate the local minima issue to some extent since this discontinuous planning problem tends to keeping same limbs for supporting, which will lead to local minimas given different settings, \eg, limited planning round. Two safety factors for toe slip and motor over-torque are considered in the planner. Instead of tuning randomly, they are interpretable. The results show that a variety of transition contact sequences is generated in order to handle different environment setups, including slippery wall surfaces and steps/inclines that can be taken advantage of. Hardware demonstrations verify the feasibility and show that our planner can be used for standard gait schedule of legged robots as well.

{
\bibliographystyle{ieeetr}
\bibliography{references}
}

\end{document}